\documentclass{article}

\usepackage[final,nonatbib]{nips_2017}

\usepackage[utf8]{inputenc} 
\usepackage[T1]{fontenc}    
\usepackage{booktabs,ragged2e,amsfonts,tabularx,hyperref,url,times,
courier,csquotes,dblfloatfix,setspace,ifthen,amsmath, amsthm, amssymb, wrapfig,graphicx,amsfonts,multirow,subfig,todonotes,algorithmic,times,algorithm}

\author{
  Ershad Banijamali$^1$ , Ahmad Khajenezhad$^2$, Ali Ghodsi$^3$ , Mohammad Ghavamzadeh$^4$ 
   \\  
  $^1$School of Computer Science,  University of Waterloo \\
  $^2$Sharif University of Technology\\
  $^3$Department of Statistics and Actuarial Science, University of Waterloo \\
  $^4$DeepMind \\
  \small
  \texttt{sbanijam@uwaterloo.ca, khajenezhad@ce.sharif.edu} \\
  \small
  \texttt{aghodsib@uwaterloo.ca, ghavamza@google.com} \\
}

\title{Disentangling Dynamics and Content \\for Control and Planning}

\begin{document}

\maketitle

\begin{abstract}
In this paper, We study the problem of learning a controllable representation for high-dimensional observations of dynamical systems. Specifically, we consider a situation where there are multiple sets of observations of dynamical systems with identical underlying dynamics. Only one of these sets has information about the effect of actions on the observation and the rest are just some random observations of the system. Our goal is to utilize the information in that one set and find a representation for the other sets that can be used for planning and ling-term prediction.
\end{abstract}

\section{Introduction}

The world surrounding us is full of events that we only observe them through high-dimensional sensory data. However, in many cases, these events can be described by few features and simple relations. Discovering the simple low-dimensional feature space is an underlying task in many data processing algorithms. With the recent advances in the area of artificial neural networks, use of deep structures for learning the low-dimensional representations has been outstandingly increased in different applications.  A good representation is defined based on the task in hand. 

In the area of control, a good representation means a low-dimensional feature space, in which the relation between different states of the system can be modeled by simple functions. 
Finding such representation has been studied recently in different works \cite{bohmer2015autonomous}. Deep autoencoders have been used for obtaining an appropriate representation for control in \cite{lange2010deep,wahlstrom2015pixels}. This problem has been also studied in action respecting embedding (ARE) framework \cite{bowling2005action}. Embed to control (E2C) \cite{watter2015embed}, finds a low-dimensional locally-linear embedding of the observations that allows planning and long-term prediction by applying model predictive controllers, e.g. iterative linear quadratic regulator (iLQR). More recently, robust controllable embedding (RCE), \cite{Ershad17RCE}, has been proposed, which can handle noise in the dynamics of the system.

In this paper, we address this problem in a more generalized setting. Suppose we have different sets of high-dimensional observations from the systems that have the same underlying dynamics. Therefore, in all of the observations there exist a common set of features that correspond to the dynamics of the system. Our goal is to extract this set of features using only one set of observations and use the learned dynamics to do planning and long-term prediction for the other sets. To do so, we design a model that disentangles the features that contribute in dynamics and those who just contribute in the content of the image. Building such model requires dynamics information (i.e. knowing how the actions change our observation from the system) in one set and there is no need to have such information in other sets. 

Learning disentangled features has various applications in image and video processing and text analysis and has been studied in different works \cite{mathieu2016disentangling}. More recently, authors in \cite{tulyakov2017mocogan,denton2017unsupervised} proposed a model in the framework of generative adversarial networks (GANs) that disentangles dynamics and content for video generation. However, to the best of our knowledge, our model is the first model that proposes disentangling dynamics and content for control, planning, and prediction.

\section{Problem Statement}
Suppose we have different sets of high-dimensional observations from the states of dynamical systems where the underlying dynamics of the systems is the same. For now, let us assume that we only have one dynamical system and there are just two observation sets from this system from different angles. We make this assumption just for the sake of simplicity in notations, but it can be easily relaxed. The two observation sets are denoted by $X$ and $Y$ that belong to the observation spaces $\mathcal{X}$ and $\mathcal{Y}$, respectively.

Let us denote by $\mathcal{S}$, the true state space of the system, in which $\mathbf{s}_t$ represents the state of the system at time step $t$. The dynamics of the system in this space is defined by $f_{\mathcal{S}}$:
\begin{equation}
\mathbf{s}_{t+1} = f_{\mathcal{S}}(\mathbf{s}_t, \mathbf{u}_t) + \mathbf{n}^{\mathcal{S}}
\end{equation}
where $\mathbf{n}^{\mathcal{S}}$ is the noise in the state space.
We do not have any information about the state space and want to estimate it based on our observations.

Suppose set $X$ consists of triples $(\mathbf{x}_t,\mathbf{u}_t,\mathbf{x}_{t+1})$, i.e. observation of the system at time $t$, action that is applied to the system at time $t$, and the next observation after applying $\mathbf{u}_t$ to the system, respectively. Therefore, we know how the actions change our observations in $X$. We also assume that the observations in this set have Markov property. Set $Y$ also has some observations of the system from a different point of view. However, there is no information about the actions and the effect of the actions on our observation in this set. We denote the observations in this set by $\mathbf{y}_t$. Note that $\mathbf{x}_t$ and $\mathbf{y}_t$ are two different observations of the state $\mathbf{s}_t$.  Since $X$ and $Y$, are  observations from one system, the underlying dynamics is the same.  Suppose that our goal is to do planning and long-term prediction in $\mathcal{Y}$.  Our approach to achieve this goal is to extract the dynamics information from $X$ and leverage this information to build a model for $Y$.

\section{Model Description}
There has been some efforts in finding a representation for high-dimensional observations of dynamical systems that is suitable for planning using neural networks. Recently, Robust Controllable Embedding (RCE) \cite{Ershad17RCE} has been proposed that shows good performance on this task. The RCE model is based on introducing a graphical model for the problem that describes the relation between pairs of observations and their embedded representations. Using deep variational learning, the lower bound of the conditional distribution of the observations is maximized.

\begin{figure}[!t]
    \centering
    \subfloat[]{{\includegraphics[trim = -20mm 10mm -25mm 10mm,width=8cm]{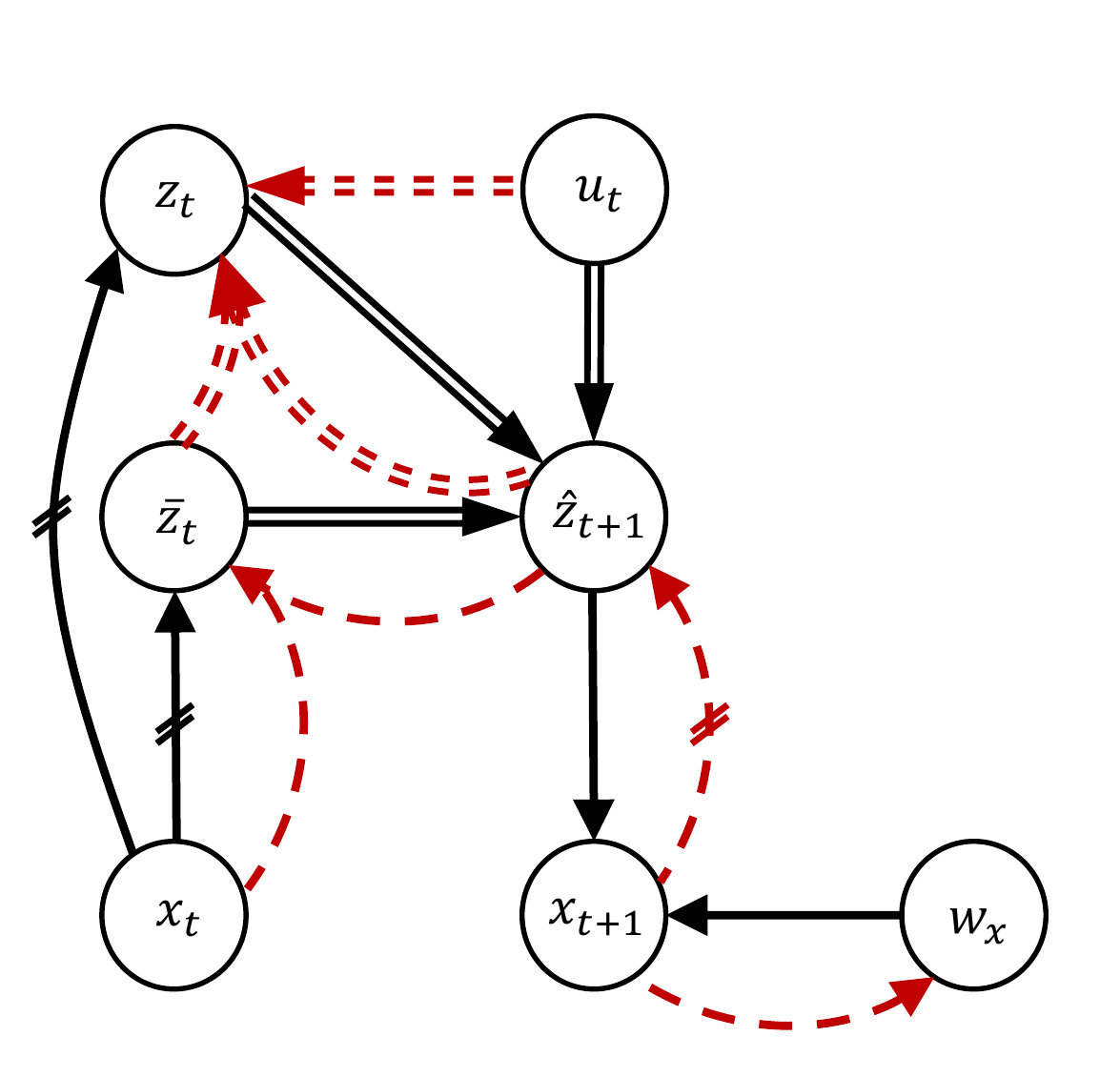} \label{fig:gm_1}}}
	\subfloat[]{{\includegraphics[trim = -25mm 20mm -20mm 10mm,width=5.5cm]{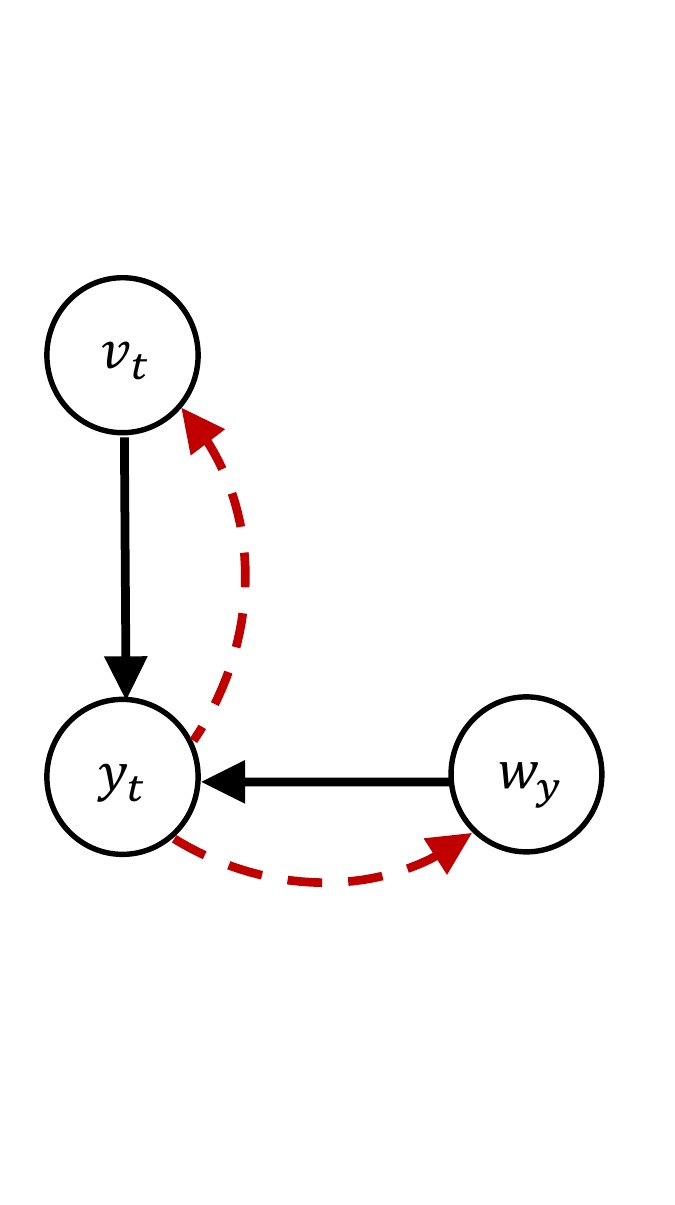} \label{fig:gm_2}}} 	
        \vspace{-.2cm}
    \caption{Graphical models. The black arrows are generative links and the red dashed ones are recognition links. The parallel lines show the deterministic links. \textbf{(a)} Graphical model for set $X$. $\bar{\mathbf{z}}_t$ and $\mathbf{z}_t$ are two samples from $p(\mathbf{z}_t|\mathbf{x}_t)$.  The neural networks that parameterize the links with hatch marks are  hard tied, i.e. $p(\mathbf{z}_t|\mathbf{x}_t) = p(\bar{\mathbf{z}}_t|\mathbf{x}_t) = q(\hat{\mathbf{z}}_t|\mathbf{x}_t)  $  . \textbf{(b)} Graphical model for $Y$ } 
    \label{fig:gm}%
    \vspace{-.3cm}    
\end{figure}

We build our model up on RCE . However, instead of using only one latent variable, we assume that there are two independent variables in the latent space. One of these variables is related to the dynamics of the system and the other one is related  to the content of the observation.  Therefore we aim to disentangle the dynamics and content in the latent space. Such disentanglement allows us to model the dynamics of the observations, even though the content of them might be very different. Consider the graphical models in Fig. \ref{fig:gm}. Fig. \ref{fig:gm_1} shows the model for $X$. In this figure, $\mathbf{z}_t$ and $\mathbf{w}_{\mathbf{x}}$ are the two latent variables that we want to represent the dynamics and content information, respectively. Similar to RCE, we want to have locally-linear dynamics in the latent space, i.e.:
\begin{equation}
\hat{\mathbf{z}}_{t+1} = \mathbf{A}_t \mathbf{z}_t+ \mathbf{B}_t \mathbf{u}_t+ \mathbf{c}_t
\label{eq:lld}
\end{equation}
where $\mathbf{A}_t$, $\mathbf{B}_t$, and $\mathbf{c}_t $ are matrices that are learned during training the model. Building this locally-linear model will allow us to use iLQR method for control. We use  $\mathbf{z}$ and $\hat{\mathbf{z}}$ to distinguish between encoding of $\mathbf{x}$ and the variable after transition. Fig. \ref{fig:gm_2} shows the model for $Y$. This set is encoded with two latent variables $\mathbf{v}_t$ and $\mathbf{w}_{\mathbf{y}}$, representing dynamics and content, respectively.  We would like to have a locally-linear dynamics similar to Eq. \ref{eq:lld}  for $\mathbf{v}$. All of the conditional distribution on these graphical models are parameterized by neural networks.

The goal in this work can be interpreted as maximizing the likelihood of observations, while imposing a further constraint that if $\mathbf{x}_t$ and $\mathbf{y}_t$ are two high-dimensional observations of the same state of the dynamical system(s), then we want $q(\mathbf{z}_t|\mathbf{x}_t)$ and $q(\mathbf{v}_t|\mathbf{y}_t)$ be close to each other, e.g. have small KL divergence. 

Suppose $q^{\star}= q(\mathbf{z}_t,\bar{\mathbf{z}}_t,\hat{\mathbf{z}}_{t+1},\mathbf{w}_{\mathbf{x}}|\mathbf{x}_t,\mathbf{x}_{t+1},\mathbf{u}_t)$ and $q^{\dagger}= q(\mathbf{v}_{t},\mathbf{w}_{\mathbf{y}}|\mathbf{y}_t)$. Based on the graphical model we can consider these factorizations for $q^{\star}$ and $q^{\dagger}$:

\begin{equation}
\begin{array}{c}
q^{\star} = q_{\phi}(\mathbf{w}_{\mathbf{x}}|\mathbf{x}_{t+1})q_{\phi}(\hat{\mathbf{z}}_{t+1}|\mathbf{x}_{t+1})q_{\varphi}(\bar{\mathbf{z}}_t|\hat{\mathbf{z}}_{t+1},\mathbf{x}_t)\delta(\mathbf{z}_t|\hat{\mathbf{z}}_{t+1},\bar{\mathbf{z}}_{t},\mathbf{u}_{t}) \\ \\ 
q^{\dagger} =  q_{\phi}(\mathbf{w}_{\mathbf{y}}|\mathbf{y}_t)q_{\phi}(\mathbf{v}|\mathbf{y}_t)

\end{array}
\end{equation}
where $\phi$ and $\varphi$ stand for encoder and transition network parameters, respectively. We also have the following factorization for the generative links in the graphical model:
\begin{equation}
p(\mathbf{x}_{t+1},\mathbf{z}_t,\bar{\mathbf{z}}_t,\hat{\mathbf{z}}_{t+1},\mathbf{w}_{\mathbf{x}}| \mathbf{x}_t,\mathbf{u}_t) = p(\bar{\mathbf{z}}_t| \mathbf{x}_t) p(\mathbf{z}_t| \mathbf{x}_t)\delta(\hat{\mathbf{z}}_{t+1}|\bar{\mathbf{z}}_t,\mathbf{z}_t,\mathbf{u}_t)p(\mathbf{x}_{t+1}|\hat{\mathbf{z}}_{t+1},\mathbf{w}_{\mathbf{x}})p(\mathbf{w}_{\mathbf{x}})
\end{equation}

In this model, we want to maximize the likelihood of all the observations. Since we consider Markov property for set $X$, maximizing the likelihood of observations in $X$ boils down to log-likelihood of the conditional distribution of the pair of observations. Therefore we will have: 
\begin{equation}
\begin{array}{l}
\log p(\mathbf{x}_{t+1}| \mathbf{x}_t,\mathbf{u}_t) + \log p(\mathbf{y}_t) \\ \\
\hspace{.8cm}\geq \mathbb{E}_{q^{\star}} \big [\log p(\mathbf{x}_{t+1},\mathbf{z}_t,\bar{\mathbf{z}}_t,\hat{\mathbf{z}}_{t+1},\mathbf{w}_{\mathbf{x}}| \mathbf{x}_t,\mathbf{u}_t)  - \log q^{\star} \big]  + \mathbb{E}_{q^{\dagger}} \big [ \log p(\mathbf{y}_t,\mathbf{v}_t,\mathbf{w}_{\mathbf{y}}) - \log q^{\dagger}\big ] \\ \\

\hspace{.8cm} = \mathbb{E}_{\substack{q_{\phi}(\hat{\mathbf{z}}_{t+1}|\mathbf{x}_{t+1})\\ q_{\phi}(\mathbf{w}_{\mathbf{x}}|\mathbf{x}_{t+1})}} \big [\log p(\mathbf{x}_{t+1}|\hat{\mathbf{z}}_{t+1},\mathbf{w}_{\mathbf{x}}) \big ] - \mathbb{E}_{q_{\phi}(\hat{\mathbf{z}}_{t+1}|\mathbf{x}_{t+1})} \big [ \text{KL} \big ( q_{\varphi}(\bar{\mathbf{z}}_t| \hat{\mathbf{z}}_{t+1},\mathbf{x}_t) \parallel p(\bar{\mathbf{z}}_t| \mathbf{x}_t)  \big ) \big ] \\ \\
\hspace{1.2cm} + \text{H} \big ( q_{\phi}(\hat{\mathbf{z}}_{t+1}| \mathbf{x}_{t+1}) \big )  + \mathbb{E}_{\substack{ q_{\phi}(\hat{\mathbf{z}}_{t+1}|\mathbf{x}_{t+1})\\q_{\varphi}(\bar{\mathbf{z}}_t|\mathbf{x}_t,\hat{\mathbf{z}}_{t+1}) } }\big [ \log p(\mathbf{z}_t| \mathbf{x}_t) \big ] - \text{KL}\big ( q_{\phi}(\mathbf{w}_{\mathbf{x}}|\mathbf{x}_t) \parallel p(\mathbf{w}_{\mathbf{x}}) \big )\\ \\
\hspace{1.2cm} + \mathbb{E}_{q^{\dagger}}\big [  \log p(\mathbf{y_t}|\mathbf{v}_t,\mathbf{w}_{\mathbf{y}}) \big ] - \text{KL}\big ( q_{\phi}(\mathbf{v}_t|\mathbf{y}_t) \parallel p(\mathbf{v}_t) \big) - \text{KL} \big ( q_{\phi}(\mathbf{w}_{\mathbf{y}}|\mathbf{y}_t) \parallel p(\mathbf{w}_{\mathbf{y}}) \big )\\ \\
\end{array}
\label{eq:RCE_obj_func}
\end{equation}

To maximize this lower bound we use the deep variational learning framework. We assume that the prior of the content variables, $\mathbf{w}_{\mathbf{x}}$ and $\mathbf{w}_{\mathbf{y}}$, are Gaussian. Also we assume $p(\bar{\mathbf{z}}_t|\mathbf{x}_t)$ is Gaussian. The constraint of minimizing the KL divergence between $q(\mathbf{z}_t|\mathbf{x}_t)$ and $q(\mathbf{v}_t|\mathbf{y}_t)$ can be imposed by considering $q(\mathbf{z}_t|\mathbf{x}_t)$ as the prior for $p(\mathbf{v}_t)$, i.e.:
\begin{equation}
p(\mathbf{v}_t) = \mathcal{N} \big ( \mu_{\phi}(\mathbf{x}_t), \mathbf{\sigma}_{\phi}(\mathbf{x}_t) \big )
\end{equation}

\begin{figure}[!t]
    \centering
\includegraphics[trim = 0mm 10mm 0mm 5mm,width=14cm]{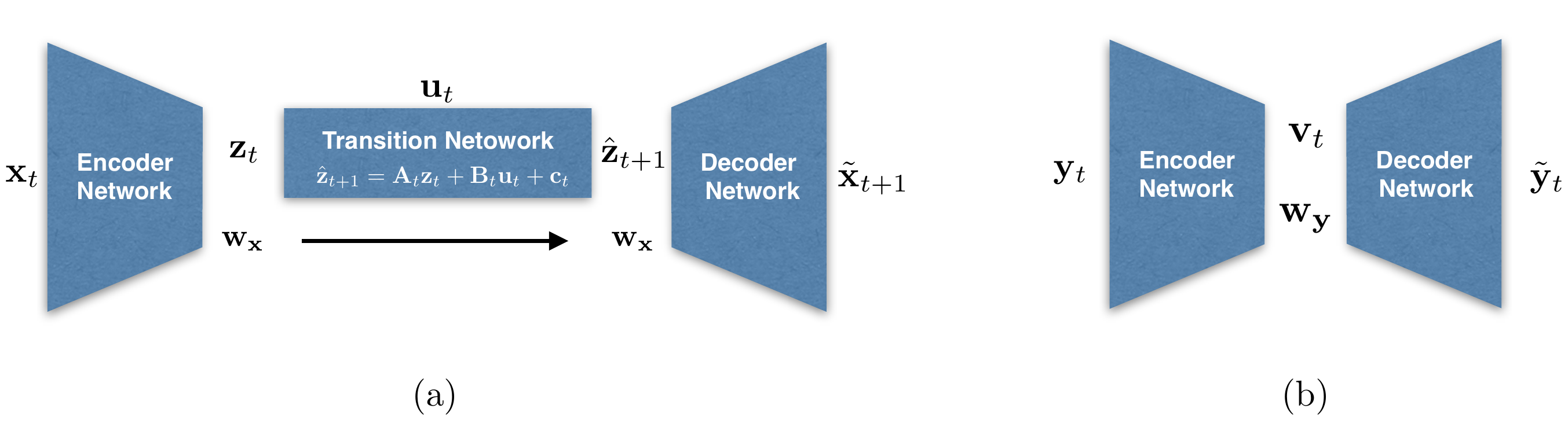} 
        \vspace{-.3cm}
    \caption{Networks of the model} 
    \label{fig:networks}%

\end{figure}

Fig. \ref{fig:networks} shows the high-level depiction of the networks in our model. In the case we use same networks for encoding and decoding the two observation sets (for example when the contents do not differ too much) , we can assume that $p(\mathbf{w}_{\mathbf{x}})$ and $p(\mathbf{w}_{\mathbf{y}})$ are two Gaussian distributions with different means. 

\section{Experiment Result}
\begin{figure}[!b]
    \centering
\includegraphics[trim = 0mm 0mm 0mm 20mm,width=13.5cm]{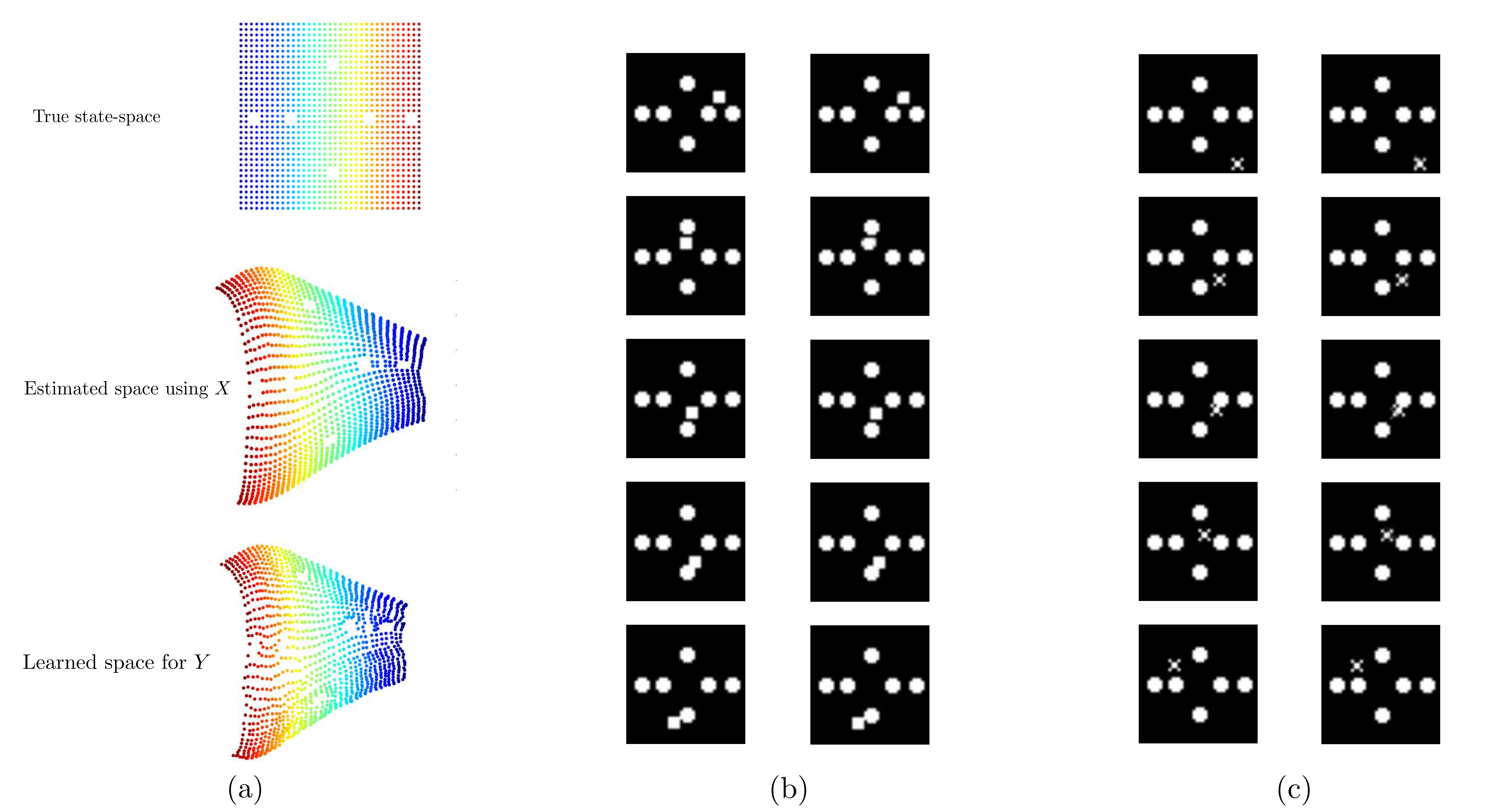} 
        \vspace{-.1cm}
    \caption{\textbf{(a)} Top: The true state space of the system. Middle: estimated locally-linear latent space from set $X$. Bottom: The hidden space learned for set $Y$. \textbf{(b)}: Left: An initial observation from $X$ on top and its next observations after applying four random actions Right: Reconstruction of the initial state and prediction of the next observations. \textbf{(c)}: Left: An initial observation from $Y$ on top and its next observations after applying four random actions Right: Reconstruction of the initial state and prediction of the next observations } 
    \label{fig:maps}%
  
\end{figure}

To evaluate the effectiveness of the proposed model, we consider the planar system domain. Consider an agent in a surrounded area, whose goal is to navigate from a corner to the opposite one, while avoiding the six obstacles in this area. The system is observed through a set of $40 \times 40$ pixel images taken from the top, which specify the agent's location in the area. Actions are two-dimensional and specify the direction of the agent's movement. Suppose that the difference between the two observation sets from this system is in the shape of the agent, as shown in Fig. \ref{fig:maps}. We use the same encoder and decoder for the two observation sets. We used $8000$ samples (triples $(\mathbf{x}_t,\mathbf{u}_t,\mathbf{x}_{t+1})$) in the set $X$ and only $2000$ samples in set $Y$.

Fig. \ref{fig:maps} shows the true map of the state-space of this system and the maps that are estimated using the model for the two observation sets. As we can see, the map that has been discovered using the information in $X$ is very well preserved for the set $Y$. In this figure we can also see some predictions of the position of the agent for both sets given some actions versus the true position of the agent after applying those action. This shows that the model is successful in learning the dynamics for $Y$ even though we did not have any information about the dynamics in this set. 

\newpage
To evaluate the performance of the model in planning, we provide different sets of initial and final observations in $\mathcal{X}$ and $\mathcal{Y}$, and use the learned models\ to find the policy that leads the agent to reach the final observation within $T$ steps. We present the performance of the model in table \ref{tbl:planar_sys} in terms of: {\bf 1)} {\em Reconstruction Loss} is the loss in reconstructing current observation  using the encoder and decoder. {\bf 2)} {\em Prediction Loss} is the loss in predicting next observations, given current observation and current action, using the encoder, decoder, and transition network. {\bf 3)} {\em Planning Loss} is computed based on the following quadratic loss:
\begin{equation}
J = \sum \limits_{t=1}^T (\mathbf{s}_t - \mathbf{s}^f)^{\top} \mathbf{Q} (\mathbf{s}_t - \mathbf{s}^f) + \mathbf{u}_t^{\top} \mathbf{R}\mathbf{u}_t.
\label{eq:traj_loss}
\end{equation}
where $\mathbf{Q}$ and $\mathbf{R}$ are cost weighting matrices. $\mathbf{s}^f$ is the state corresponding to the final observation. We apply the sequence of actions returned by iLQR to the dynamical system and report the value of the loss in Eq.~\ref{eq:traj_loss}. {\bf 4)} {\em Success Rate} shows the number of times the agents reaches the goal within the planning horizon $T$, and remains near the goal in case it reaches it in less than $T$ steps. For each of the sets, all the results are averaged over $20$ runs.

\begin{table*}[!t]
\caption{ Planar System}
\vspace{-.25cm}
\small
\begin{center}
\begin{tabular}{ c | c c c c}
 \textbf{Dataset} & \textbf{Reconstruction Loss} & \textbf{Prediction Loss} & \textbf{Planning Loss} & \textbf{Success Rate} \\
\hline
\textbf{with action ($X$) }	 &  $3.6 \pm 1.7$ & $6.2 \pm 2.8$  & $21.4  \pm 2.9$ & $100 \% $\\
\textbf{without action ($Y$)}	 & $3.9 \pm 2.2$ & $6.3 \pm 3.0$ & $22.0 \pm 2.4$	 & $100 \%$ 
\end{tabular}
\label{tbl:planar_sys}
\end{center}
\vspace{-.5cm}
\end{table*}


\vspace{-.2cm}
\section{Discussion}
\vspace{-.1cm}
This model has potential applications in self-driving cars. Self-driving cars use many sensors to observe the surrounding environment that includes expensive sensors for dynamics estimation. They also use multiple cameras to monitor the area.  Observations from the camera are rich in term of information about the content (objects in the area), however, extracting dynamics information using these observations is a hard task. On the other hand, the dynamics estimator sensors are poor in terms of the content information but provide information about action-state space with high accuracy. If we can find a way to transfer the learned dynamics from the sensor to the cameras, we can remove the sensor at the test time and reduce the cost of experiments.

 \vspace{-.2cm}
\small
\bibliography{DDC}
\bibliographystyle{abbrv}

\end{document}